\documentclass{article}

\usepackage{arxiv}
\usepackage{tabularx}
\usepackage[ruled,linesnumbered,commentsnumbered]{algorithm2e}
\usepackage[utf8]{inputenc}

\usepackage[bookmarks=false]{hyperref}
\usepackage{listings}
\usepackage{color}
\usepackage{graphicx}%
\usepackage[utf8]{inputenc} 
\usepackage[T1]{fontenc}    
\usepackage{url}            
\usepackage{booktabs}       
\usepackage{amsfonts}       
\usepackage{nicefrac}       
\usepackage{microtype}      
\usepackage{lipsum}
\usepackage{mathptmx}
\usepackage{hyperref}
\usepackage{graphicx}
\usepackage{subcaption} 
\usepackage{footnote}
\usepackage{amsmath}
\usepackage{amssymb}
\usepackage{mathrsfs}
\usepackage{array}
\usepackage{cite}
\usepackage{multirow}
\usepackage{siunitx}
\usepackage{tablefootnote}
\usepackage{mathtools}
\usepackage{commath}
\usepackage{scalerel}
\usepackage[utf8]{inputenc}
\title{Cause-Effect Preservation and Classification using Neurochaos Learning}
\author{
Harikrishnan N B$^{a,b}$, Aditi Kathpalia$^{c}$ Nithin Nagaraj$^{b}$\\
$^a$The University of Trans-Disciplinary Health Sciences And Technology \\
Bengaluru, 560 064, Karnataka, India\\
$^b$Consciousness Studies Programme,\\ National Institute of Advanced Studies,\\ Indian Institute of Science Campus, Bengaluru, 560 012, Karnataka, India. \\  
$^c$Department of Complex Systems,\\ Institute of Computer Science of the Czech Academy of Sciences,\\
Prague, Czech Republic \\
\texttt{harikrishnannb@nias.res.in, kathpalia@cs.cas.cz, nithin@nias.res.in  } \\
}
\begin{document}
\maketitle
\begin{abstract}
Discovering cause-effect from observational data is an important but challenging problem in science and engineering. In this work, a recently proposed brain inspired learning algorithm namely-\emph{Neurochaos Learning} (NL) is used for the classification of cause-effect from simulated data. The data instances used are generated from coupled AR processes, coupled 1D chaotic skew tent maps, coupled 1D chaotic logistic maps and a real-world prey-predator system. The proposed method consistently outperforms a five layer Deep Neural Network architecture for coupling coefficient values ranging from $0.1$ to $0.7$.  Further, we investigate the preservation of causality in the feature extracted space of NL using Granger Causality (GC) for coupled AR processes and and Compression-Complexity Causality (CCC) for coupled chaotic systems and real-world prey-predator dataset. This ability of NL to preserve causality under a chaotic transformation and successfully classify cause and effect time series (including a transfer learning scenario) is highly desirable in causal machine learning applications.  
\end{abstract}

{\bf Keywords:~} Neurochaos Learning, Granger Causality, Compression-Complexity Causality, Coupled Auto Regressive Processes, Coupled Chaotic Maps, Causal Machine Learning, Transfer Learning
\section{Introduction}
\label{sec:Intro}

Understanding causal factors responsible for the occurrence of a phenomena is vital for the progress of research in science and technology. The development of drugs/ vaccines to combat epidemic from time to time requires to pass the test of trustworthiness. This can be done by understanding the casual influence of the drug in combating the epidemic. The gold standard method used for this test is the \emph{Randomized Control Trials} (RCT). RCTs are used  in clinical research to discover cause-effect relation via clinical interventions. However, RCTs are not a universal answer~\cite{graz2007beyond_myth_rct} and suffer from their own challenges~\cite{nichol2010challenging_rct, concato2004observational, gautama2021rcts}. In many cases, RCTs are not technically viable, time consuming, expensive and even unethical. In such a scenario, the need to develop novel causal inference and causal discovery algorithms to learn from \emph{observational data}~\cite{mooij2016distinguishing} plays a crucial role. The availability of high computational resources and big data motivates the  development of data driven learning algorithms. Machine Learning (ML) and Deep Learning (DL) owes to the abundance of data and computational resources for its growth. Despite the success of ML/DL algorithms in the field of natural language processing, computer vision, speech recognition, these algorithms face difficulty in interpretability and trustworthiness. This is due to the fact that these algorithms are merely discovering associations between `input' and `output' in the name of `learning'. However, discovering associations alone is insufficient as an explanation to aid in the decision making process. Often, what is needed is a {\it causal} explanation which goes over and beyond associations (or correlations). Hence, arises the need for causal machine learning algorithms where causality meets learning. 

Judea Pearl's ladder of causation provides a three level hierarchical order of a casual learner~\cite{pearl2018book}. The three levels proposed are: (1) Association, (2) Intervention and (3) Counterfactuals.  Most of the current causality testing methods are at the level of association. Some of the popular causality testing methods for time series data are Granger Causality (GC)~\cite{granger1969investigating}, Transfer Entropy (TE)~\cite{schreiber2000measuring_TE}, data compression based causality testing method namely Compression-Complexity Causality~\cite{kathpalia2019data}. GC and TE are popular causality testing methods used in econometrics~\cite{hiemstra1994testing, chiou2008economic}, climatology~\cite{mosedale2006granger, stips2016causal}, neuroscience~\cite{seth2015granger, vicente2011transfer} etc. However, both GC and TE lie at the lowest rung of the ladder - {\it associational} causality. CCC is an example of {\it interventional} causality measure and corresponds to the second rung of this ladder.

In this research, we use a recently proposed brain inspired learning algorithm namely Neurochaos Learning~\cite{harikrishnan2021noise} (NL) to learn generalized causal patterns from time series data. NL draws its inspiration from the chaotic firing of neurons in the brain~\cite{korn2003there_chaos_II}. We investigate the following: (a) can NL learn unique patterns of cause and effect from observational data?, (b) do the chaos based features extracted from the input layer of NL preserve causality? In order to tackle the first question, a binary classification problem is formulated for the classification of cause-effect. The efficacy of NL and DL in learning generalized cause-effect pattern from data and thereby performing classification tasks are investigated. The second question is addressed by studying the preservation of cause-effect for the features extracted from the input layer of NL. To this end, we use Granger Causality (GC) and Compression-Complexity Causality (CCC). 

The sections in the paper are arranged as follows: Section 2 describes the proposed method used to do the cause-effect classification. Section 3 provides details of the simulated data used to carry out the experiments. Section 4 deals with experiments, results and discussions on simulated and real world prey-predator dataset, as well as a transfer learning scenario. Section 5 addresses the limitations and scope for future work. The concluding remarks are provided in Section 6.

\section{Proposed Method}

Neurochaos Learning (NL) is a novel brain inspired neuronal learning algorithm that has been recently proposed. The authors in ~\cite{balakrishnan2019chaosnet,harikrishnan2020neurochaos} claim that NL is inspired from the chaotic firings of neurons in the brain and has mainly two architectures: (a) \verb+ChaosNet+~\cite{harikrishnan2020neurochaos}, (b) ChaosFEX+ML. In~\cite{laleh2020chaotic}, the authors employ \verb+ChaosNet+ for continual learning.  In another work~\cite{chen2021deep}, the authors propose deep \verb+ChaosNet+ for action recognition in videos.

Inspired by these recent developments, in this work,  we employ \verb+ChaosNet+ architecture for the classification of cause-effect from observational data. The architecture consists of an input layer of Generalized L\"{u}roth Series (GLS) neurons which are one-dimensional (1D) chaotic skew tent maps described as follows:
\begin{eqnarray}
    T(x)=
    \begin{cases}
      \frac{x}{b}, &  0 \leq x < b,\\
      \frac{(1-x)}{(1 - b)}, & b \leq x < 1,
    \end{cases}
    \label{eq_skew_tent}
  \end{eqnarray}
where the skewness of the map is controlled by the parameter $b$ ($0<b<1$). Upon arrival of the input data/ stimulus $(x_k)$, the chaotic GLS neurons in the input layer starts firing (from the initial value $q$) until the chaotic neural trace of the neurons reaches the $\epsilon$ neighbourhood of the corresponding stimulus. From the neural trace thus generated, the following features are extracted:
\begin{enumerate}

    \item \emph{Firing time ($N$)}: The amount of time the chaotic neural trace takes to recognise the input stimulus. 
    
    \item \emph{Firing rate ($R$)}: Fraction of time the chaotic neural trace is above the discrimination threshold $b$ so as to recognize the stimulus.
        \item \emph{Energy ($E$)}: For the chaotic neural trace $y(t)$ with firing time $N$, energy is defined as: 
        \begin{equation}
            E = \sum_{t = 1}^N |y(t)|^{2}.
        \end{equation}
    \item \emph{Entropy ($H$)}: For the chaotic neural trace $y(t)$, we first compute the binary symbolic sequence $Sym(t)$ as follows:
\begin{eqnarray}
Sym(t_i) = \left\{\begin{matrix}
 0, &  y(t_i) < b,  \\
 1, & b \leq y(t_i) < 1,\\
\end{matrix}\right.
\end{eqnarray}

where $i = 1$ to $N$ (firing time). We then compute Shannon Entropy of $Sym(t)$ as follows:
    \begin{equation}
        H(Sym) = -\sum_{i = 1}^2 p_i \log_2(p_i) ~~\text{bits},
    \end{equation}
    where $p_1$ and $p_2$ refers to the probabilities of the symbols $0$ and $1$ occurring in $Sym(t)$ respectively. 
\end{enumerate}
For each input value $s$ (stimulus) of a data instance of class $k$ is mapped to a 4D vector $[N_s, R_s, E_s, H_s] $. The collection of these 4D vectors forms the ChaosFEX feature space. If the input data is 1D with $Z$ classes, the mean
representation vector of the $k$-th class is given by $\frac{1}{m} [\sum_{i = 1}^{m} N_i, \sum_{i = 1}^{m} R_i, \sum_{i = 1}^{m} E_i, \sum_{i = 1}^{m} H_i ]$ where $m$ is the number of data instances (training data) for each class. 

%

In the case of \verb+ChaosNet+, the classifier computes the cosine similarity of the ChaosFEX features extracted from the test sample with the pre-computed mean representation vectors (consisting of mean values of ChaosFEX features for each stimuli) of each class. The predicted class is assigned the label corresponding to the maximum cosine similarity. A detailed explanation of the \verb+ChaosNet+ architecture and its working is provided in~\cite{balakrishnan2019chaosnet}. 

\section{Datasets}
To evaluate the efficacy of \verb+ChaosNet+ and deep learning for the classification of cause-effect, we used simulated datasets from (a) Coupled autoregressive (AR) processes, (b) Coupled 1D chaotic maps in master-slave configuration (1D skew tent maps and the 1D logistic maps) and real-world dataset from a (c) prey-predator system.

\subsection{Coupled AR processes}
The governing equations for the coupled AR processes are the following:
\begin{eqnarray}
M(t) = a_{1}M(t-1) + \gamma r(t),\\
S(t) = a_{2}S(t-1) + \eta M(t-1) + \gamma r(t),
\label{eq_master_slave_coupled_AR}
\end{eqnarray}
where $M(t)$ and $S(t)$ are the independent and the dependent time series respectively; $a_1 = 0.8$, $a_2 = 0.9$,  the noise intensity $\gamma = 0.03$ and $r(t)$ is the i.i.d additive gaussian noise drawn from a standard normal distribution. The coupling-coefficient $\eta$ is varied from $0$ to $1$ in steps of $0.1$. We generated $1000$ independent random trials for each value of $\eta$. Each of the data instances are of length $2000$, after removing the initial $500$ samples (transients) from the time series.

\subsection{Coupled 1D Chaotic maps in Master-Slave configuration}

\subsubsection{Coupled Skew-tent maps}
\begin{figure}[!h]
    \centerline{ \includegraphics[width=0.47\textwidth]{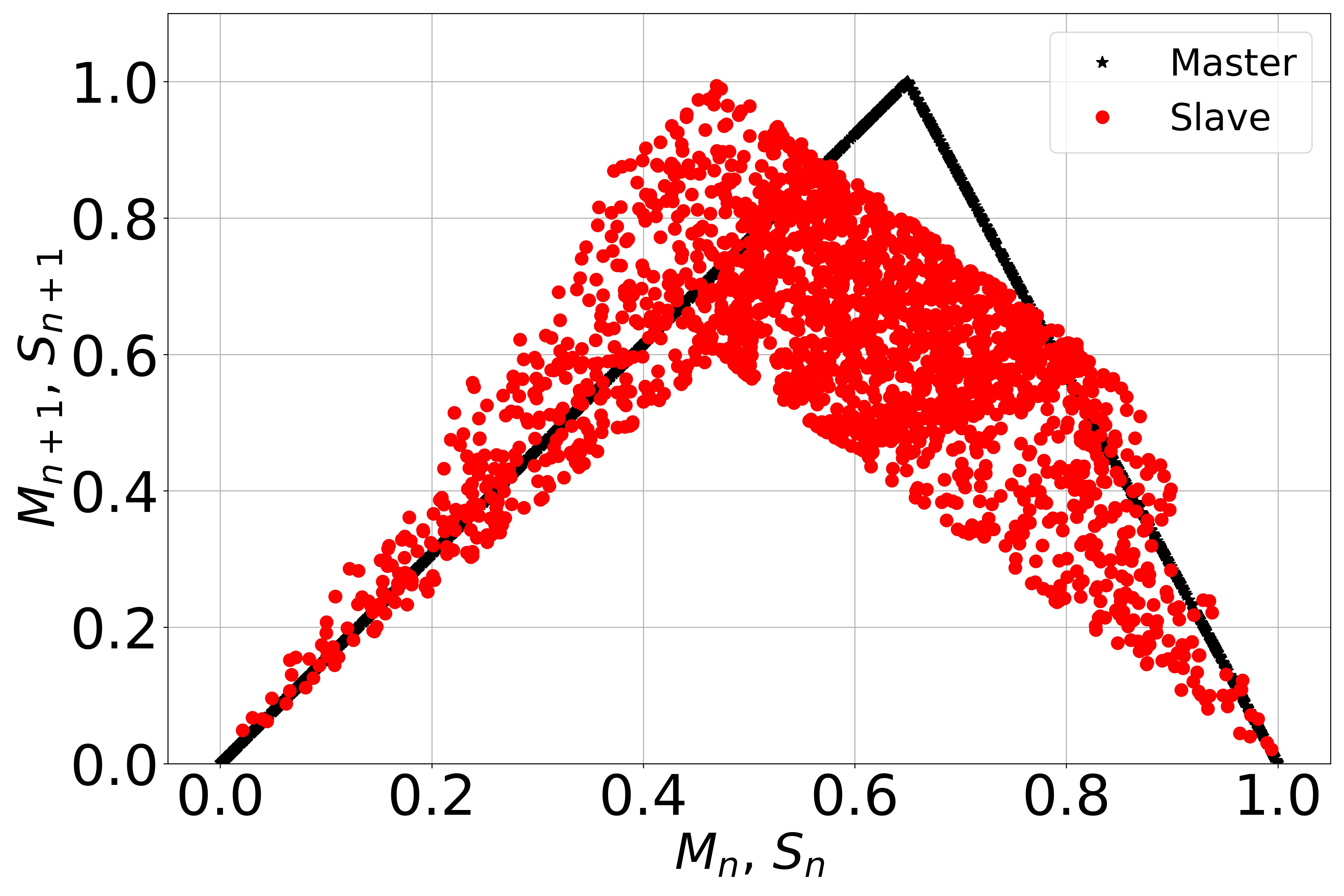}}
    \caption{Attractor for the coupled 1D chaotic skew tent maps in master slave configuration with $b_1 = 0.65$ and $b_2 = 0.47$.}
    \label{Fig_attractor_train_coupled_skew_tent}
    \end{figure}

The governing equations used to generate the master and slave time series for the coupled 1D skew-tent maps are the following: 
\begin{eqnarray}
    M(n)=T_{1}(M(n -1)),\\
    S(n) = (1-\eta)T_{2}(S(n-1)) + \eta M(n-1),
    \label{eq_master_slave_coupled_tent_map}
\end{eqnarray}

where $M(n)$ is the master and $S(n)$ is the slave system. $M(n)$ influences the dynamics of the slave system (equation~\ref{eq_master_slave_coupled_tent_map}). The coupling coefficient is given by $\eta$ is varied from $0$ to $0.9$ with a step size of $0.1$. $T_{1}(n)$, and $T_{2}(n)$ are skew tent maps with skewness $b_1 = 0.65$, and $b_2 = 0.47$ respectively. The initial values are chosen randomly for the master-slave system in the interval $(0,1)$. We generated $1000$ independent random trials for each value of $\eta$. Each of the data instances are of length $2000$, after removing the initial $500$ samples (transients) from the time series. As an example, the attractor for this master-slave system for $\eta = 0.4$ is provided in Figure~\ref{Fig_attractor_train_coupled_skew_tent}. 

\subsubsection{Coupled Logistic maps}
The 1D Logistic map is a widely used model to study population dynamics~\cite{may2004simple}.  The governing dynamics for coupled logistic maps in master-slave configuration is given by:
\begin{eqnarray}
    M(n)=L_{1}(M(n-1)),\\
    S(n) = (1-\eta)L_{2}(S(n-1)) + \eta M(n-1),
\end{eqnarray}
The coupling coefficient $\eta$ is varied from $0$ to $0.9$.  $L_{1}(n) = 4 \cdot L_{1}(n-1)(1-L_{1}(n-1))$, and  $L_{2}(n) = 3.82 \cdot L_{2}(n-1)(1-L_{2}(n-1))$. The attractor for this coupled dynamical system is provided in Figure~\ref{Fig_Case_V_attractor_logistic_map}.

For both systems, $1000$ data instances ($M(n), S(n)$) are generated and grouped as class-0 ($M(n)$: Cause) and class-1 ($S(n)$: Effect) respectively. Table~\ref{table_train_test_distri_chaotic_skew_tent_map} gives details of the train-test split for the classification task.
\begin{table}[!h]
\centering
\caption{Train-Test distribution for the simulated datasets.\label{table_train_test_distri_chaotic_skew_tent_map}}
\begin{tabular}{|c|c|c|}
\hline
Class   & Traindata & Testdata \\ \hline
Class-0 & 801       & 199      \\ \hline
Class-1 & 799       & 201      \\ \hline
Total   & 1600      & 400      \\ \hline
\end{tabular}
\end{table}

\section{Experiments, Results and Discussions}
In this section, we begin with a description of hyperparameter tuning for NL and DL followed by a demonstration of causality preservation by ChaosFEX for coupled AR processes (and the failure of DL). Subsequently, macro F1-scores for \verb+ChaosNet+ and DL for the cause-effect classification for each $\eta$ are plotted. For all results in this paper, software implementation is performed using Python 3, scikit-learn~\cite{scikit-learn}, keras~\cite{chollet2015keras}, ChaosFEX toolbox~\cite{harikrishnan2021noise}, Multivariate Granger Causality (MVGC) toolbox~\cite{barnett2014mvgc}, CCC toolbox~\cite{kathpalia2019data} and MATLAB.
\subsection{Hyperparameter tuning for NL}

Every ML algorithm has a set of hyperparameters that needs to be tuned for efficient performance. In the case of \verb+ChaosNet+, there are three hyperparameters - initial neural activity ($q$), discrimination threshold ($b$), and noise intensity ($\epsilon$)~\cite{harikrishnan2021noise}. The hyperparameter tuning is done only once with the traindata corresponding to $\eta = 0.4$ (Table~\ref{table_train_test_distri_chaotic_skew_tent_map}) separately for the coupled AR processes and coupled skew tent maps. 

For a fixed value of $b = 0.499$, and $\epsilon = 0.171$, $q$ was varied from $0.01$ to $0.98$ with a stepsize of $0.01$ for both coupled AR processes and coupled chaotic skew tent maps. In the case of coupled AR processes, a maximum average macro F1-score $= 0.605$ is obtained for $q = 0.78$. In the case of coupled skew tent maps, a maximum average macro F1-score = $1.0$ was obtained for the following values of $q =$ [0.16, 0.26, 0.27, 0.28, 0.29, 0.30, 0.31, 0.32, 0.34, 0.36, 0.37, 0.38, 0.48, 0.51, 0.52, 0.56, 0.57, 0.72, 0.76, 0.77, 0.78, 0.79, 0.81, 0.82, 0.83, 0.84, 0.85, 0.86, 0.87, 0.88, 0.91, 0.92, 0.93, 0.94, 0.95, 0.96, 0.98] in a five-fold cross validation using traindata. We choose $q = 0.56$ for further experiments.

\subsection{Deep Learning Parameters}

A five layer Deep Learning architecture was used to evaluate the efficacy of cause effect classification. The number of nodes in the input layer $= 2000$, followed by first hidden layer with $5000$ neurons and sigmoid activation function. The output from this layer is passed to second hidden layer with $500$ neurons and ReLU activation function. This is followed by $100$ neurons with Relu activation function in the third hidden layer. The fourth hidden layer contains $30$ neurons with ReLu activation function. The output layer contains $2$ neurons with softmax activation function. Training was done for $30$ epochs.
\subsection{Preservation of Granger Causality for coupled AR processes under a chaotic transformation}
\begin{figure}[!h]
    \centerline{ \includegraphics[width=0.47\textwidth]{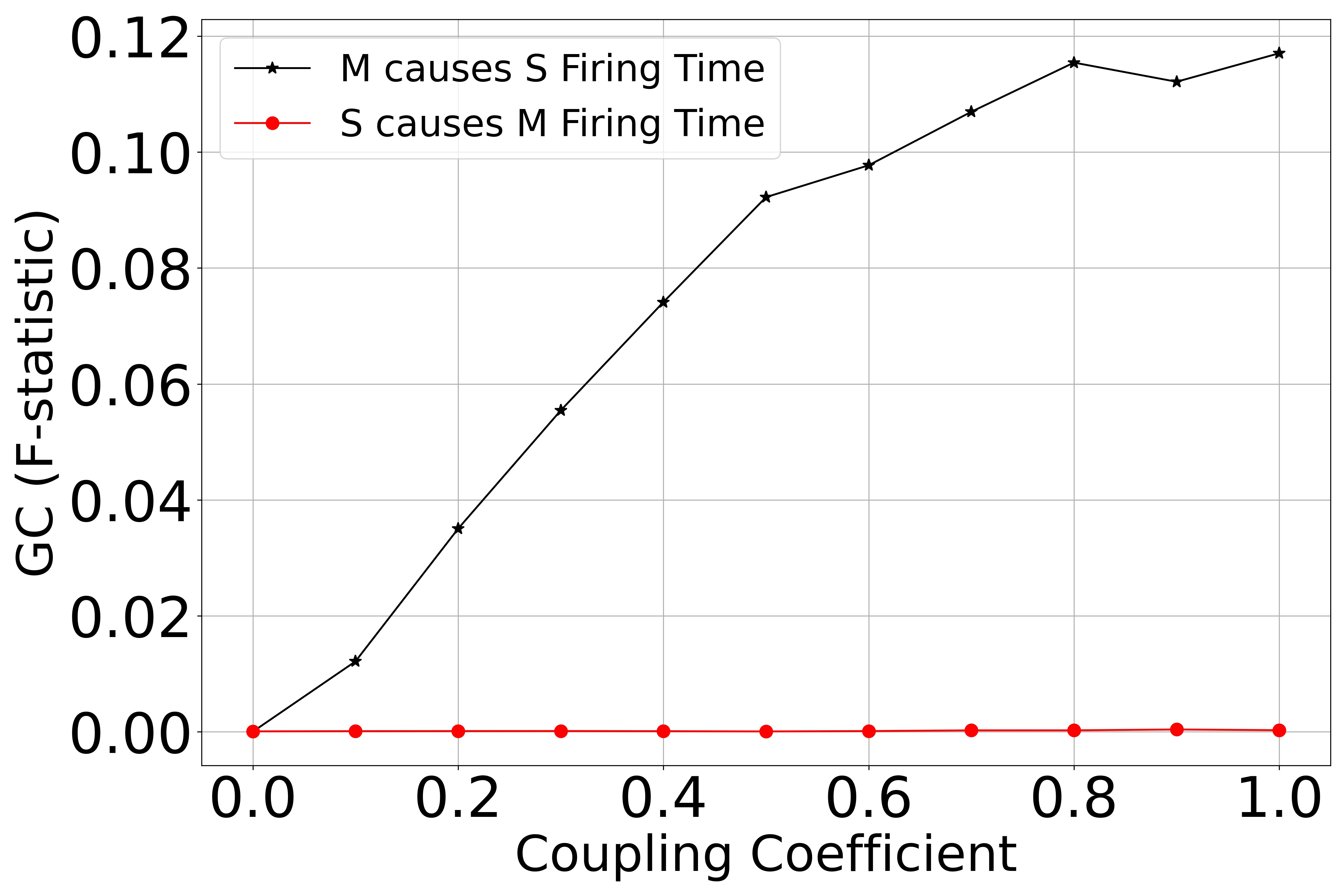}}
    \caption{GC vs. coupling coefficient for the firing time feature extracted from the coupled AR processes. The ChaosFEX settings are $q = 0.78$, $b = 0.499$, and $\epsilon = 0.171$. The GC F-statistic is computed from $50$ trials.}
    \label{Fig_GC_vs_coupling_coefficient_AR_process}
    \end{figure}

\begin{figure}[!h]
    \centerline{ \includegraphics[width=0.47\textwidth]{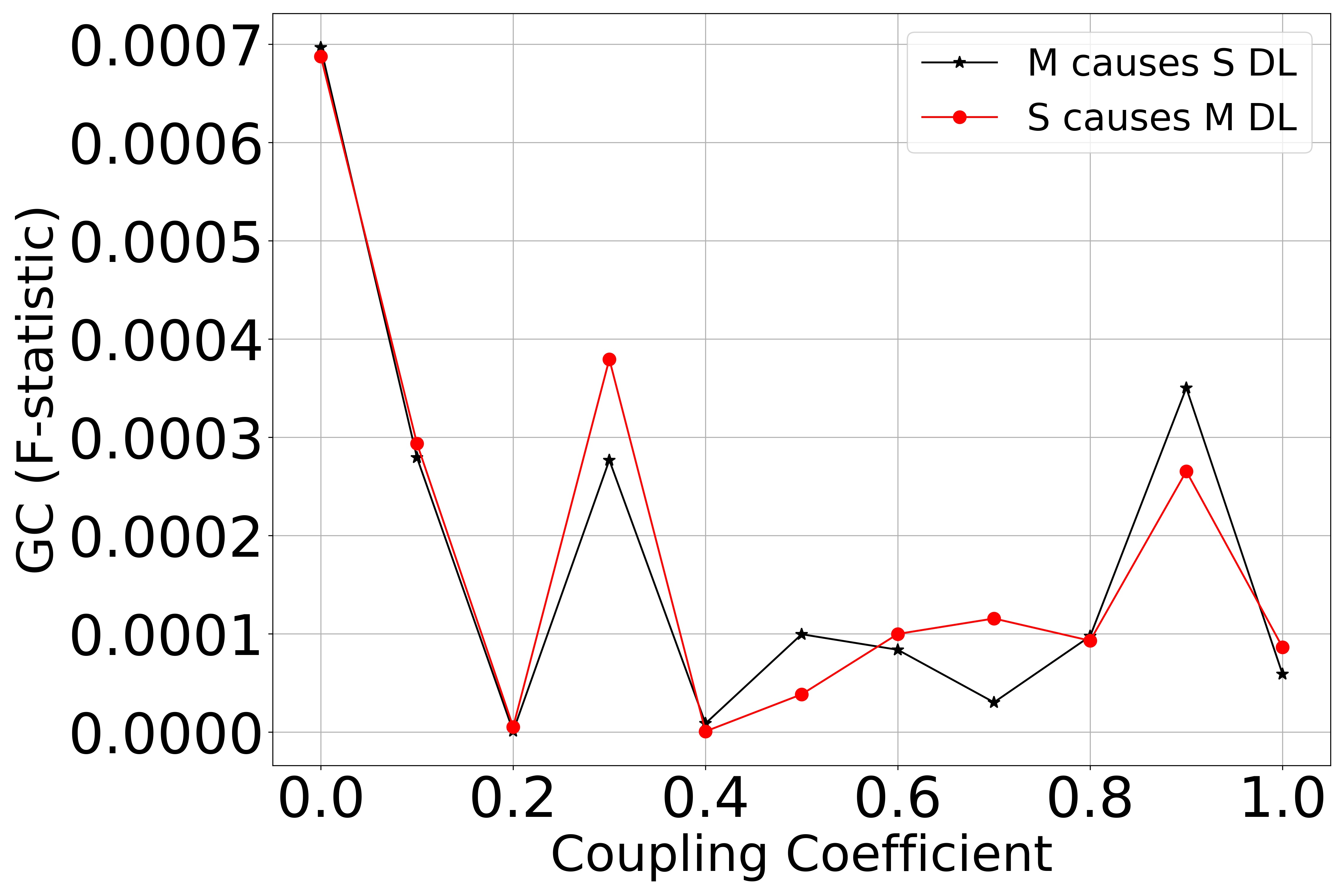}}
    \caption{GC vs. coupling coefficient for DL features extracted from the fourth hidden layer of a five layer neural network. The GC F-statistic is computed from $50$ trials.}
    \label{Fig_GC_vs_coupling_coefficient_AR_process_DL}
    \end{figure}

Accurate estimation of causality for coupled AR processes is ideally suited for Granger Causality (GC) since GC models time series as AR processes. This is the reason GC is very popular in causal analysis of financial time series, climatology and neuroscience. ChaosFEX features are extracted after a chaotic transformation of the input time series. It is important to verify whether Granger Causality is preserved under such a nonlinear transformation. To test this, we perform the following experiment. For $q = 0.78$, $b = 0.499$, and $\epsilon = 0.171$, the firing time has been extracted from ChaosFEX for time series from coupled AR processes. The GC vs. coupling coefficient plot for firing time depicted in Figure~\ref{Fig_GC_vs_coupling_coefficient_AR_process} reveals that indeed Granger Causality is nicely preserved. The GC values shown here are obtained from 50 random trials\footnote{The maximum model order setting in the MVGC toolbox was set to $30$ for ChaosFEX features and to $20$ for DL features.}. This indicates the reliability of the chaotic transformation of NL in preserving granger causality and hence very desirable in applications which employ GC. Note that such a property is not available for DL (Figure~\ref{Fig_GC_vs_coupling_coefficient_AR_process_DL}) making NL a very attractive candidate for causal ML applications. 

\subsection{Classification of Cause Effect for Coupled Skew-Tent maps in Master-Slave Configuration}
\begin{figure}[h!]
    \centerline{ \includegraphics[width=0.47\textwidth]{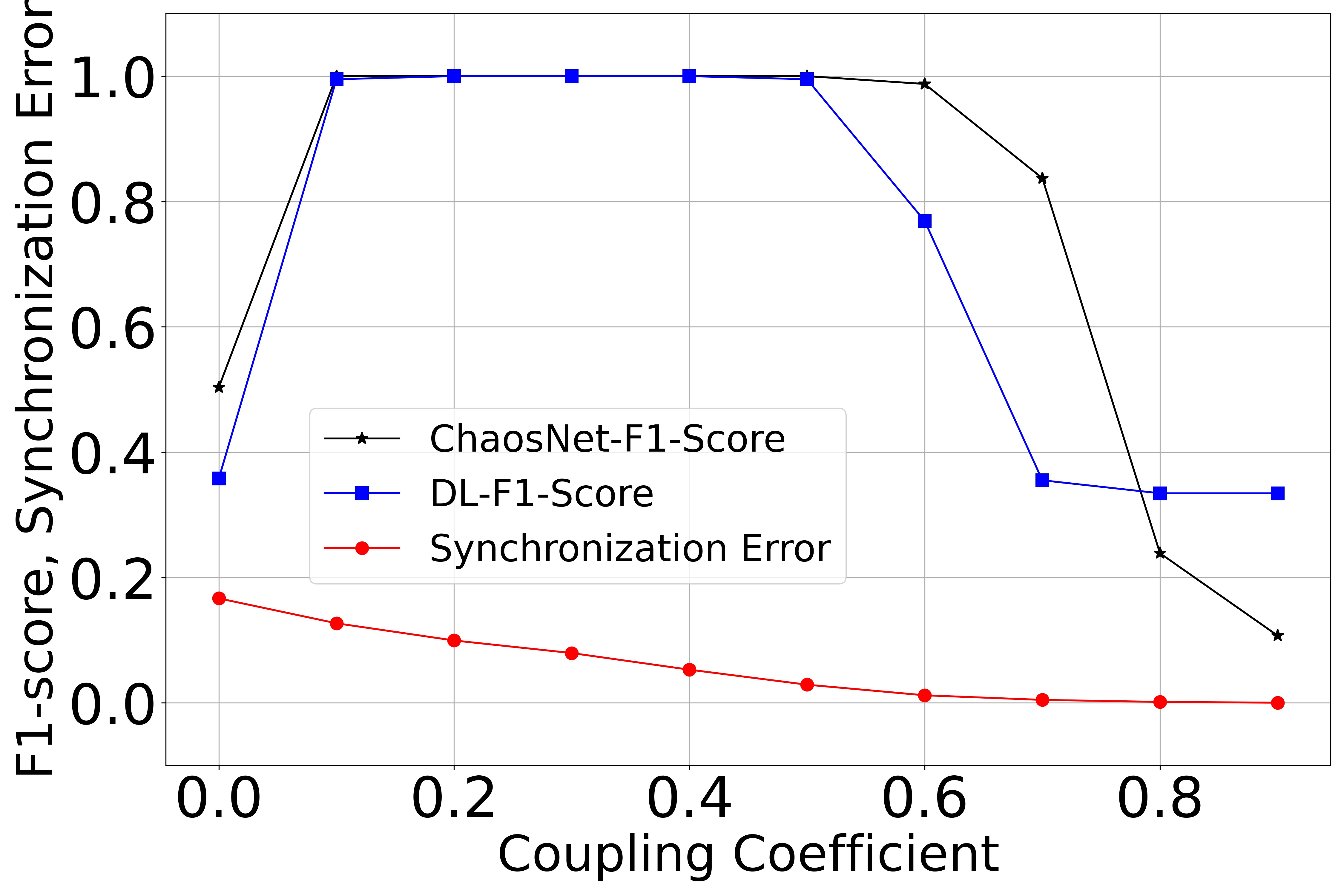}}
    \caption{Performance comparison of ChaosNet and five layer DL for the classification of cause-effect for 1D coupled skew tent map in master-slave configuration.}
    \label{Fig_chaosnet_dl_skew_tent_coupled_data}
    \end{figure}

The performance of \verb+ChaosNet+ and 5 layer DL for varying coupling coefficient $(\eta)$ is depicted in Figure~\ref{Fig_chaosnet_dl_skew_tent_coupled_data}.

\verb+ChaosNet+ and DL give identical performance (a macro F1-score $= 1.0$) for $\eta$ values up to  $0.5$. However, for $\eta =[0.6, 0.7]$, \verb+ChaosNet+ outperforms DL. Beyond $\eta > 0.6$, the synchronization error $< 0.013$ indicating that the two time series are practically identical. Hence, classification fails as there is essentially nothing to distinguish between the two time series owing to synchronization.

\subsection{Preservation of causality in ChaosFEX feature space}

\begin{figure}[!h]
    \centerline{ \includegraphics[width=0.47\textwidth]{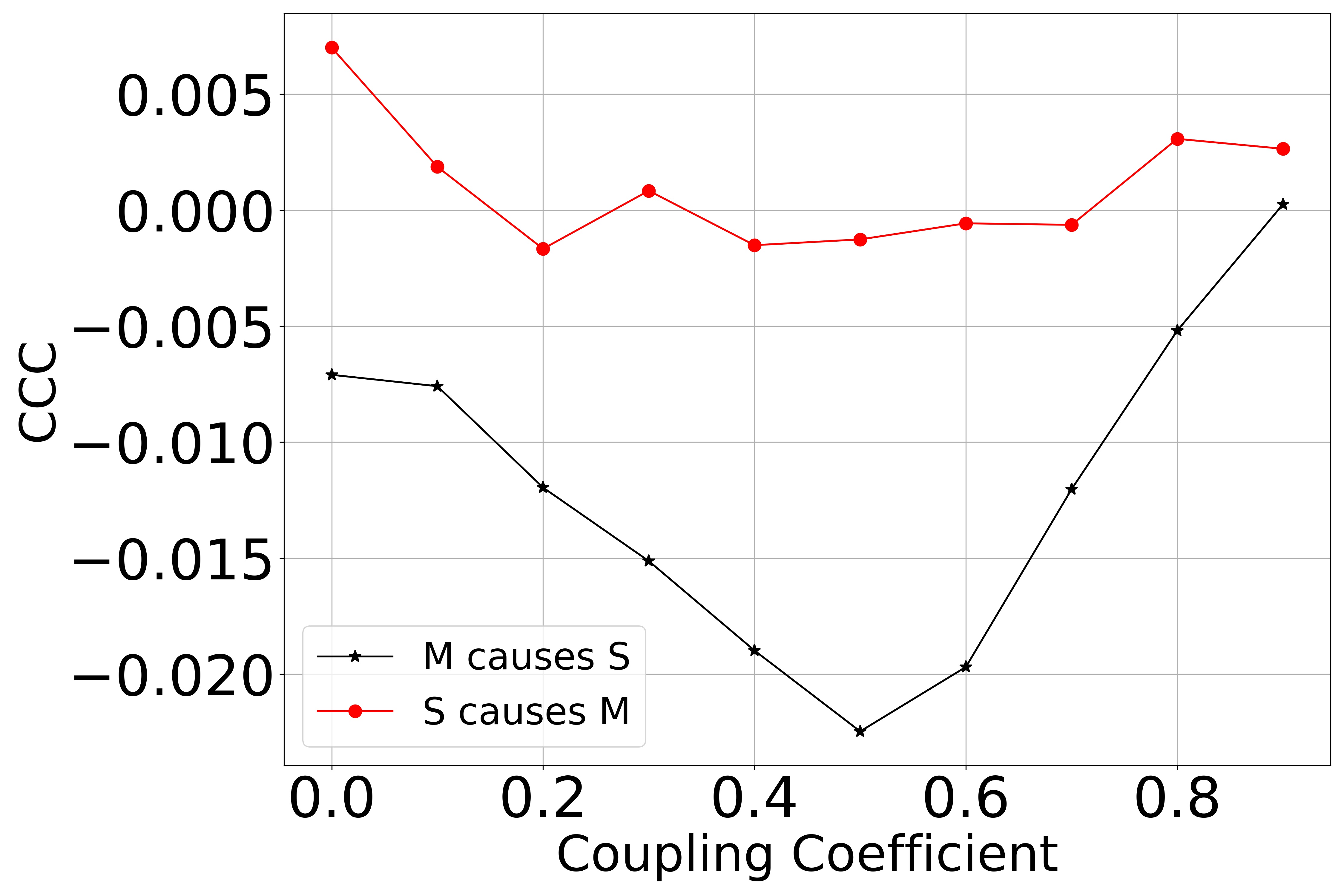}}
    \caption{CCC vs Coupling Coefficient for the raw data corresponding to 1D chaotic skew tent map in master-slave configuration.}
    \label{Fig_ccc_vs_coupling_coeff_skew_tent_coupled_raw_data}
    \end{figure}

To check if causality is preserved in the ChaosFEX feature space of unidirectionally coupled skew-tent maps, we use the measure Compression-Complexity Causality (CCC)~\cite{kathpalia2019data}. CCC is ideal for application to non-linear time series, where often GC can face issues. Figure~\ref{Fig_ccc_vs_coupling_coeff_skew_tent_coupled_raw_data} shows the CCC estimates for original (raw) time series. The estimates plotted are averaged over $50$ trials with CCC parameters\footnote{These were chosen using the selection criteria described in~\cite{kathpalia2019data}.} set to $L=100, w=15, \delta=50, B=4$. As expected, the magnitude of CCC values from the master to the slave increases with increasing coupling and begins to decrease as the time-series become synchronized and effectively no transfer of information can be detected. As discussed in~\cite{kathpalia2019data}, CCC can take negative values and its magnitude denotes the strength of causation. CCC values in the direction of causation from slave to master are much lower in magnitude and remain close to zero.

\begin{figure}[!h]
    \centerline{ \includegraphics[width=0.47\textwidth]{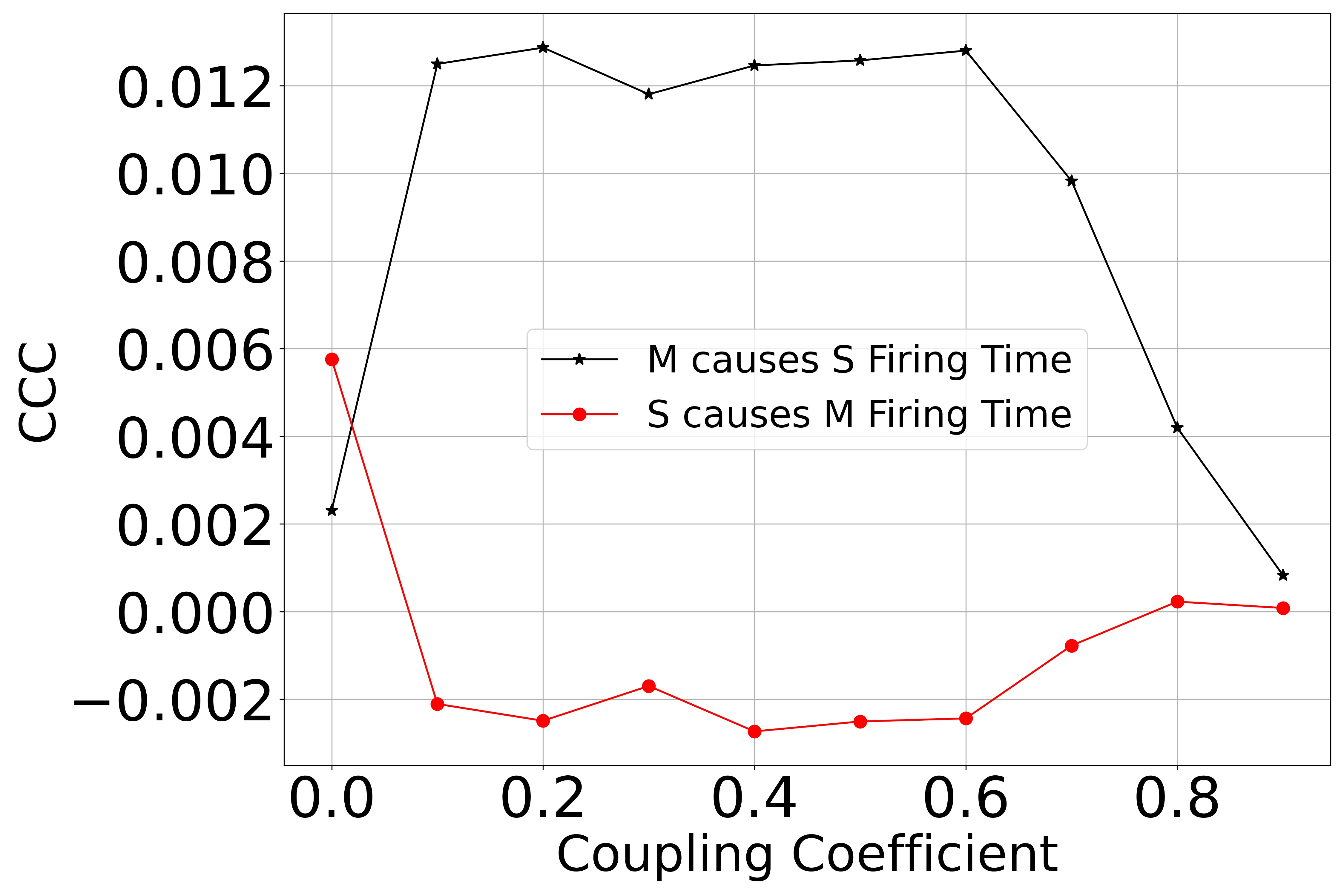}}
    \caption{CCC vs Coupling Coefficient for firing time (ChaosFEX feature) corresponding to 1D chaotic coupled skew tent maps in master-slave configuration.}
    \label{Fig_ccc_vs_coupling_coeff_skew_tent_coupled_data}
    \end{figure}

CCC for the corresponding firing time feature of ChaosFEX for these coupled maps is depicted in Figure~\ref{Fig_ccc_vs_coupling_coeff_skew_tent_coupled_data}. These values are also averaged over $50$ trials and computed with CCC parameters set to $L=120, w=15, \delta=60, B=2$. Here, master to slave CCC does not perfectly preserve the increasing trend with increasing values of coupling, however decreases just as the estimates for raw data, when the processes proceed to synchronization. The slave to master estimates for the coupling range $0.1-0.9$ are quite low in magnitude and remain close to zero as expected. Even though the estimates for zero coupling are not very close to zero or take exactly the same value and the increasing trend for increasing coupling is not perfectly preserved, CCC estimates in the direction for which coupling exists and the its opposite are well differentiated and hence it can be said that ChaosFEX features do a reasonably good job in preserving causality even for skew chaotic tent maps. Surrogate based causality analysis might help to reveal a more adequate picture of the differentiation and of the existence of causality, but is out of the scope of this work.

%
%
%
\newpage
\subsection{Transfer Learning for Cause-Effect Classification}
We have demonstrated the possibility of cause-effect classification for coupled chaotic maps in master-slave configuration. However, it is interesting to explore whether we can transfer this `learning' to scenarios where the master-slave systems are different from the ones for which the method was trained. Specifically, we shall change the skewness of both the master and slave systems from the original parameter values used in the training phase. A more drastic case of transfer learning would be to test on an entirely different nonlinear map, for example, coupled logistic maps without training afresh (using the same learned parameters as the coupled skew tent maps). These would help us determine to what extent the learning is generalizable for both NL and DL. 


We consider the following cases for transfer causal learning:
\begin{itemize}
    \item \textbf{Case I:} Train with master-slave coupled skew tent map system ($b_1 = 0.65$ , $b_2 = 0.47$) and test with master-slave coupled skew tent map system with $b_1 = 0.6$ and $b_2 = 0.4$ (classification results are in Figure~\ref{Fig_Case_I_manifold_learning}). The attractor for skew tent map master slave testdata with $b_1 = 0.6$ and $b_2 = 0.4$ is provided in Figure~\ref{Fig_Case_I_attractor_skew_tent}.
     \begin{figure}[!h]
    \centerline{ \includegraphics[width=0.47\textwidth]{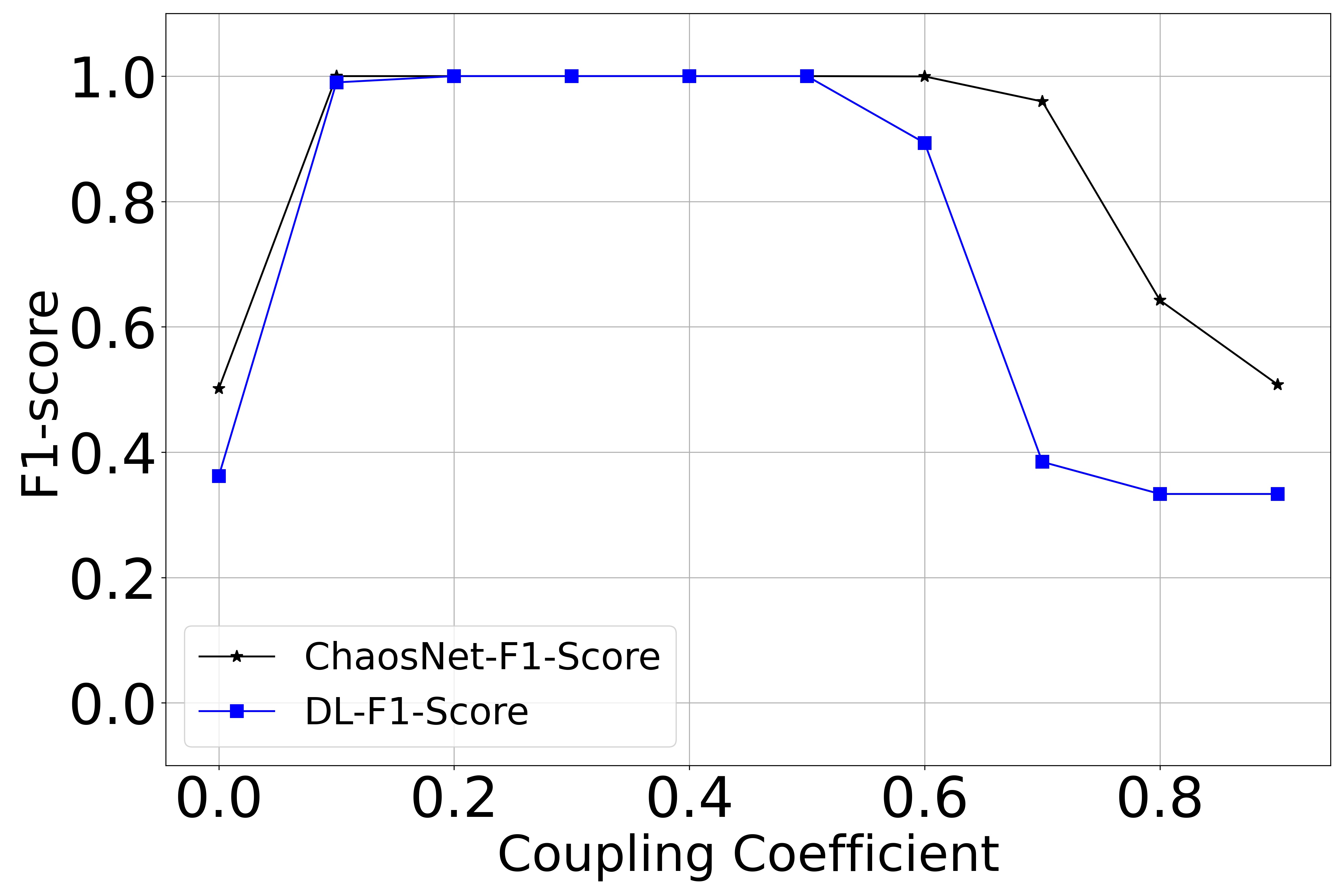}}
    \caption{Transfer learning for case I: comparative performance of ChaosNet and five layer DL evaluated using macro F1-score for $\eta$ in the range $0$ to $0.9$ with a stepsize of $0.1$.}
    \label{Fig_Case_I_manifold_learning}
    \end{figure}
    
         \begin{figure}[!h]
    \centerline{ \includegraphics[width=0.47\textwidth]{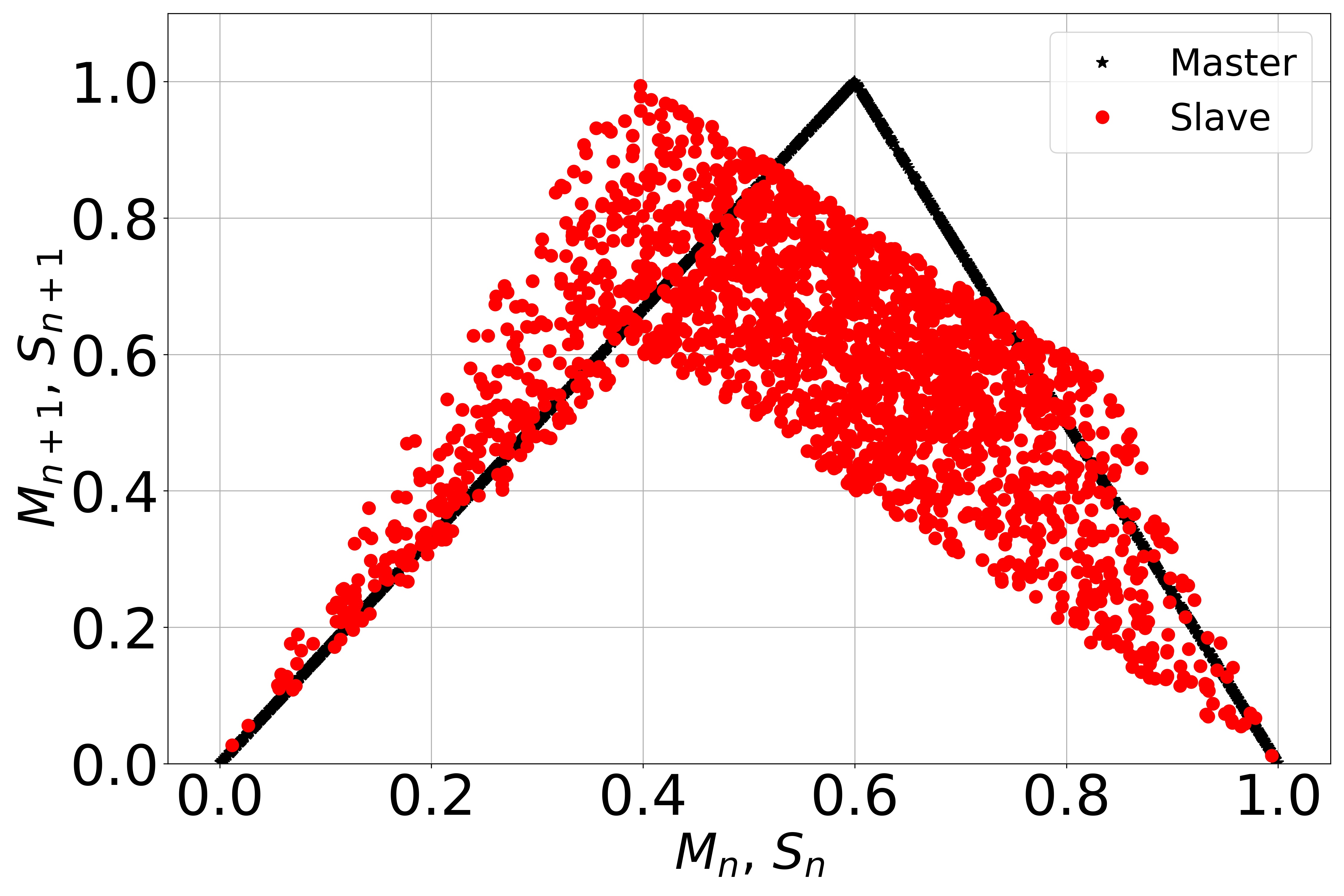}}
    \caption{Case I: Attractor for the coupled 1D chaotic skew tent maps in master slave configuration with $b_1 = 0.6$ and $b_2 = 0.4$.}
    \label{Fig_Case_I_attractor_skew_tent}
    \end{figure}
    
    \item \textbf{Case II:} Train with master-slave coupled skew tent map system ($b_1 = 0.65$, $b_2 = 0.47$) and test with master-slave coupled skew tent map system with $b_1 = 0.1$ and $b_2 = 0.3$ (classification results are in Figure~\ref{Fig_Case_II_manifold_learning}). The attractor for skew tent map master slave testdata with $b_1 = 0.1$ and $b_2 = 0.3$ is provided in Figure~\ref{Fig_Case_II_attractor_skew_tent}.
         \begin{figure}[!h]
    \centerline{ \includegraphics[width=0.47\textwidth]{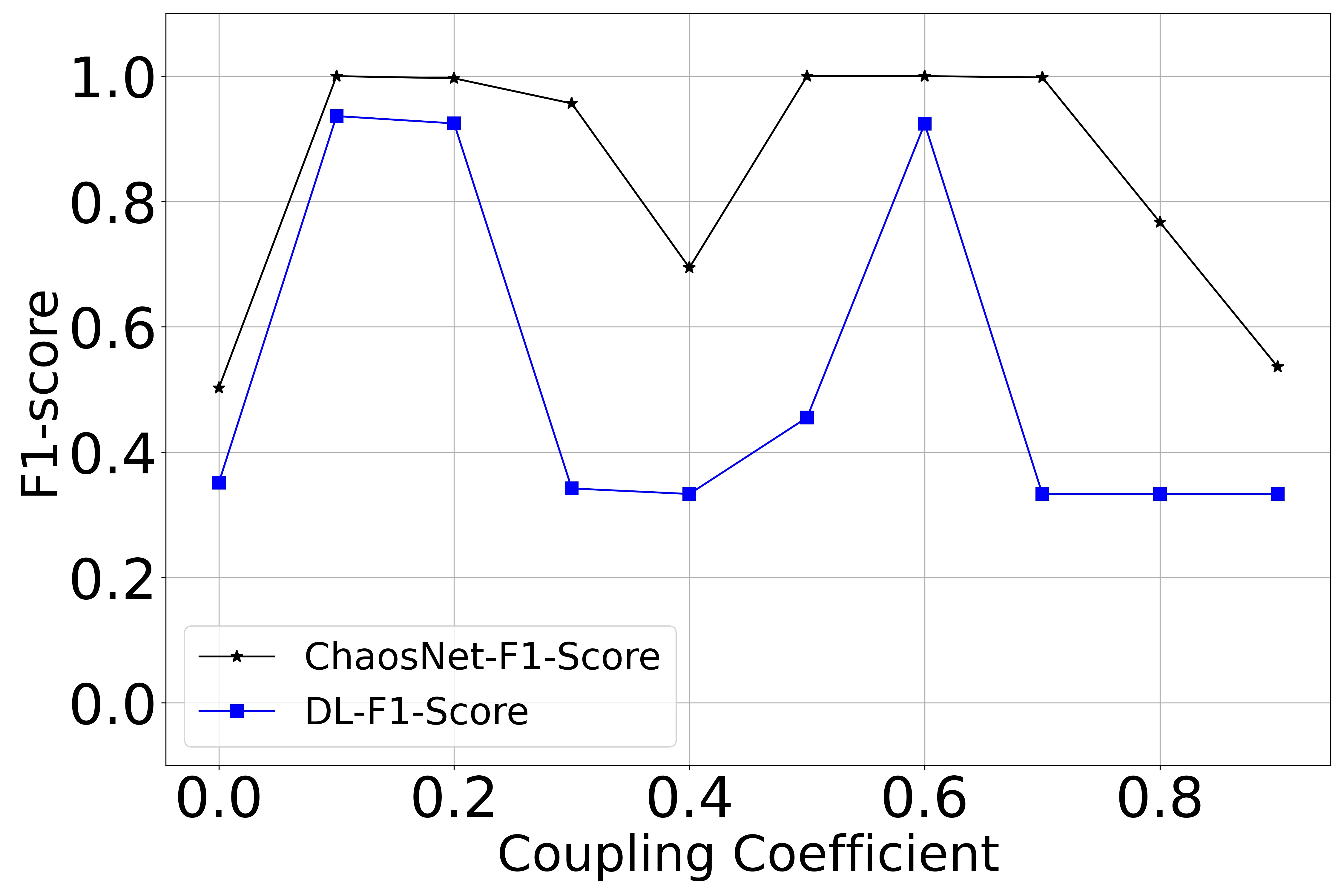}}
    \caption{Transfer Learning for Case II: comparative performance of ChaosNet and five layer DL evaluated using macro F1-score for $\eta$ in the range $0$ to $0.9$ with a stepsize of $0.1$.}
    \label{Fig_Case_II_manifold_learning}
    \end{figure}
    
             \begin{figure}[!h]
    \centerline{ \includegraphics[width=0.47\textwidth]{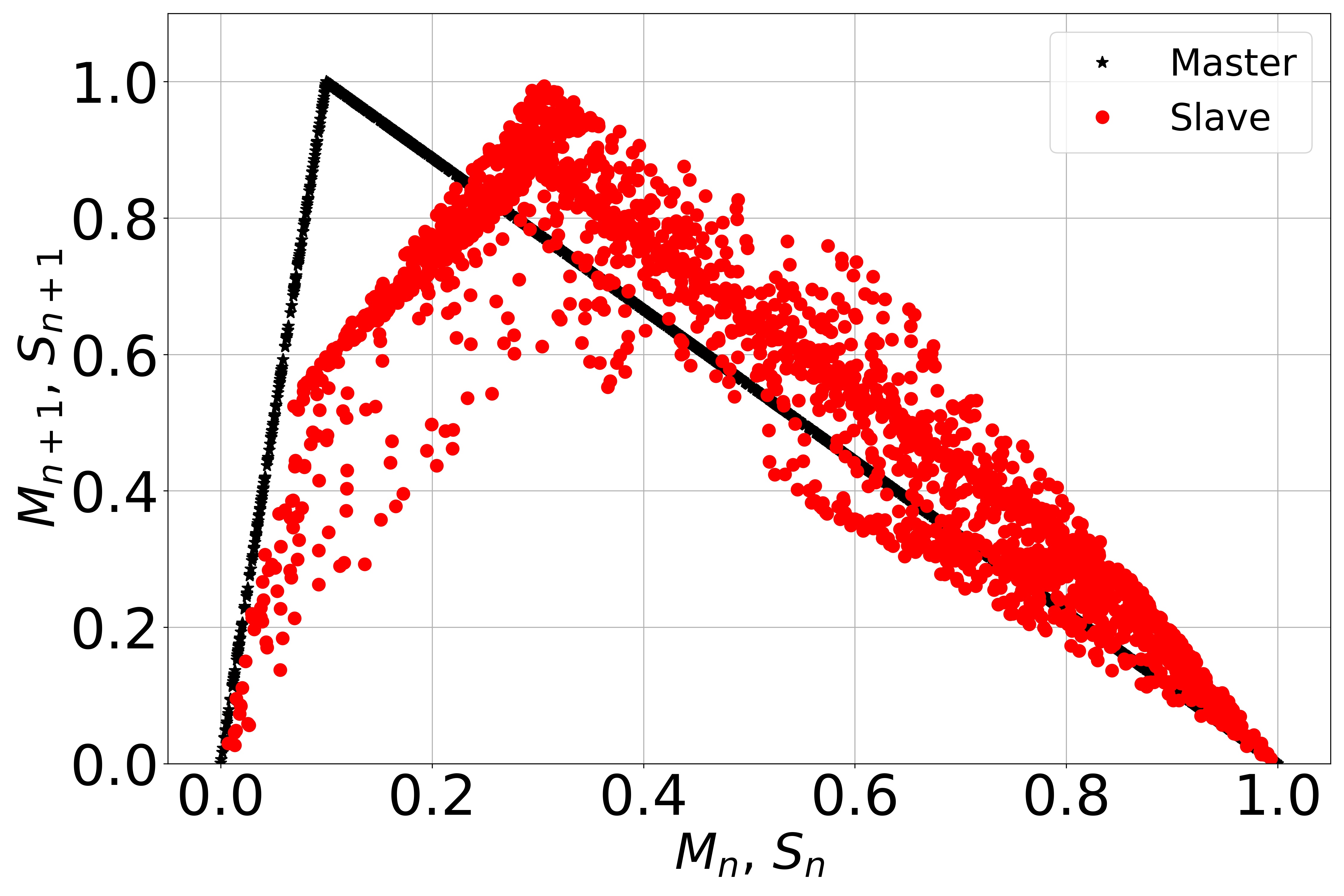}}
    \caption{Case II: Attractor for the coupled 1D chaotic skew tent maps in master slave configuration with $b_1 = 0.1$ and $b_2 = 0.3$.}
    \label{Fig_Case_II_attractor_skew_tent}
    \end{figure}

    \item \textbf{Case III:} Train with master-slave coupled skew tent map system ($b_1 = 0.65$, $b_2 = 0.47$) and test with master-slave coupled skew tent map system with $b_1 = 0.49$ and $b_2 = 0.52$ (classification results are in Figure~\ref{Fig_Case_III_manifold_learning}). The attractor for skew tent map master slave testdata with $b_1 = 0.49$ and $b_2 = 0.52$ is provided in Figure~\ref{Fig_Case_III_attractor_skew_tent}.
         \begin{figure}[!h]
    \centerline{ \includegraphics[width=0.47\textwidth]{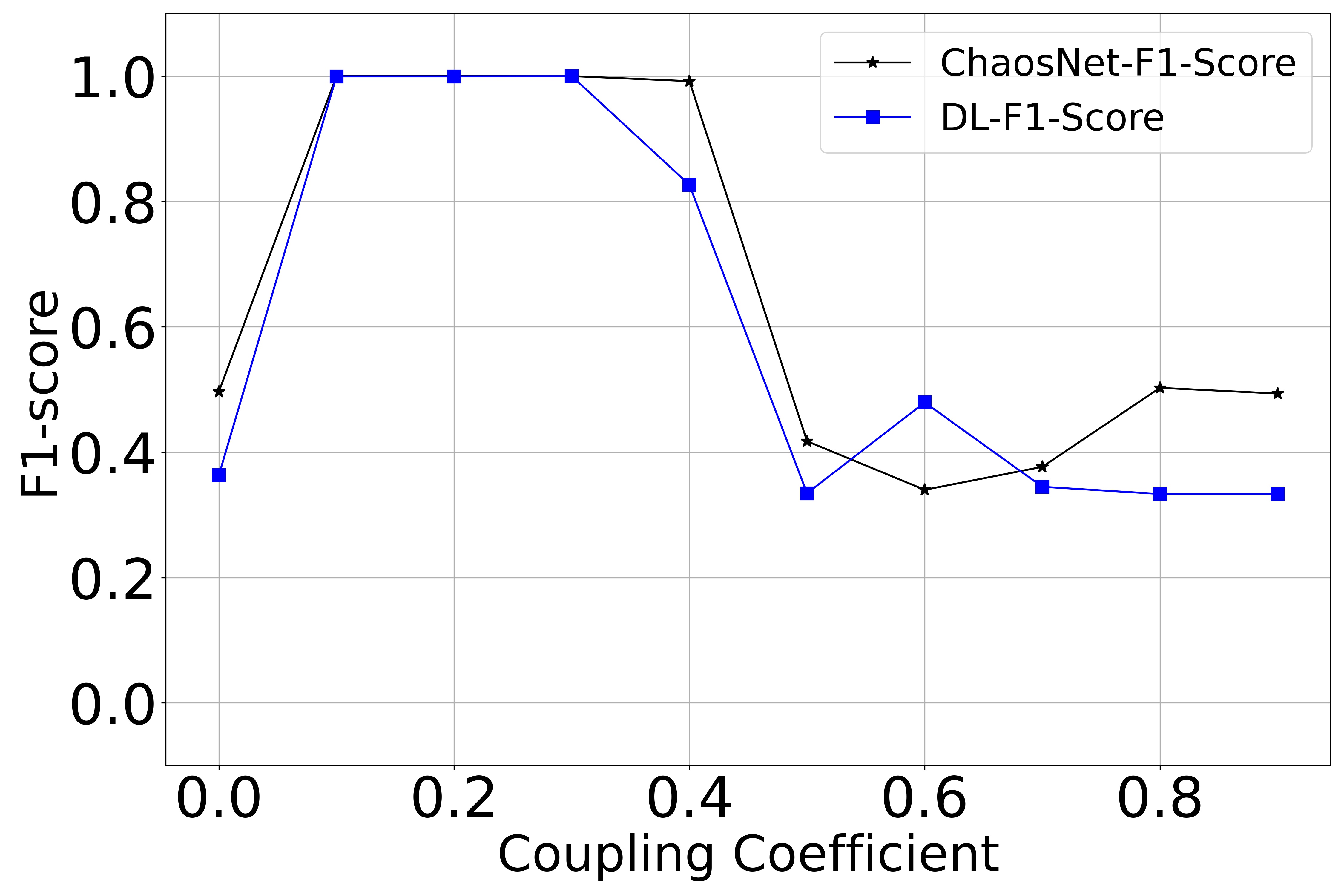}}
    \caption{Transfer Learning for Case III: comparative performance of ChaosNet and five layer DL evaluated using macro F1-score for $\eta$ in the range $0$ to $0.9$ with a stepsize of $0.1$.}
    \label{Fig_Case_III_manifold_learning}
    \end{figure}
    
                 \begin{figure}[!h]
    \centerline{ \includegraphics[width=0.47\textwidth]{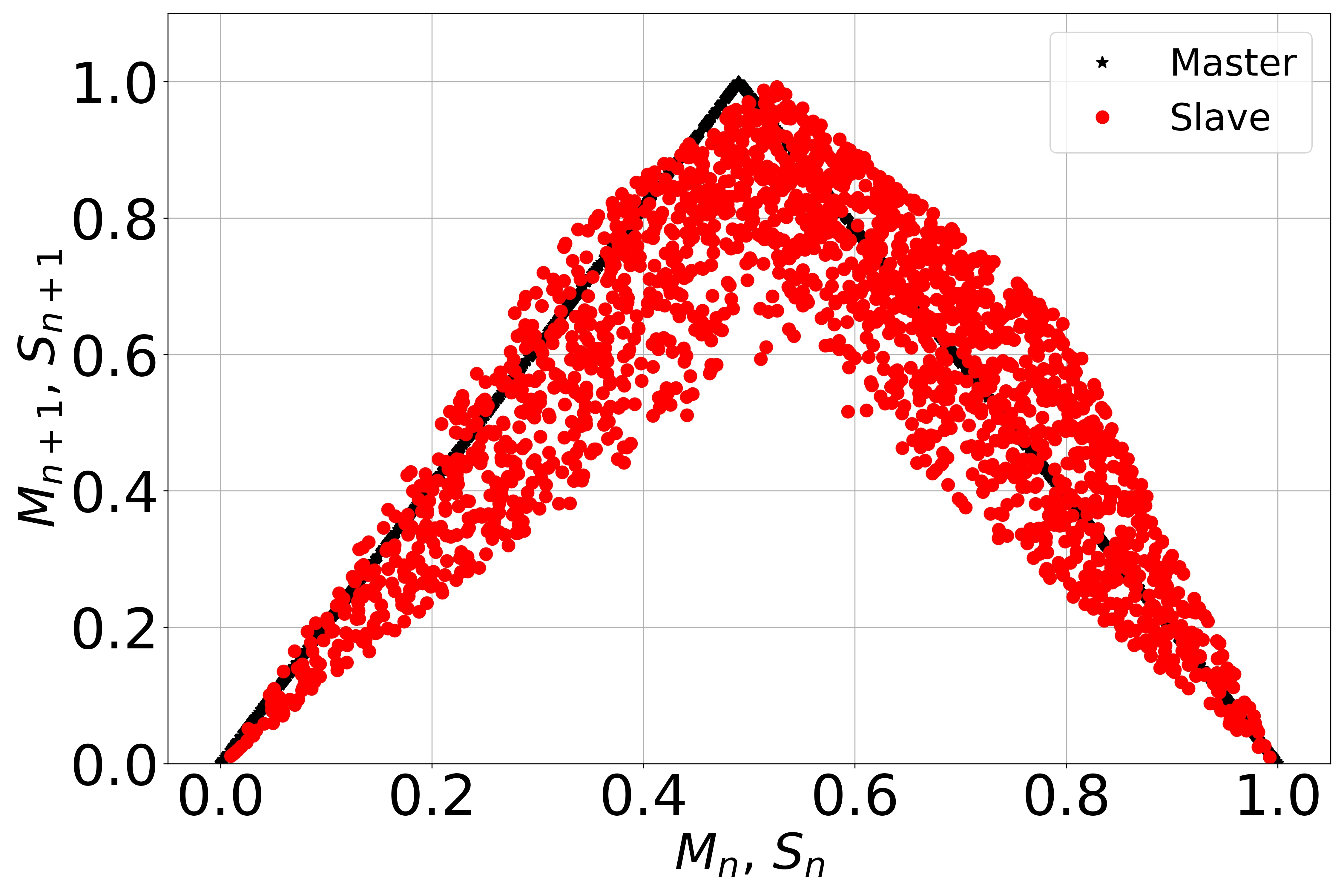}}
    \caption{Case III: Attractor for the coupled 1D chaotic skew tent maps in master slave configuration with $b_1 = 0.49$ and $b_2 = 0.52$.}
    \label{Fig_Case_III_attractor_skew_tent}
    \end{figure}
    
    
 \item \textbf{Case IV:} Train with skew tent map master-slave coupled skew tent map system ($b_1 = 0.65$, $b_2 = 0.47$) and test with logistic map master-slave system with $A_1 = 4.0$ and $A_2 = 3.82$ (Figure~\ref{Fig_Case_V_manifold_learning_logistic_map}). The attractor for logistic map master slave testdata with $A_1 = 4.0$ and $A_2 = 3.82$ is provided in Figure~\ref{Fig_Case_V_attractor_logistic_map}.
      \begin{figure}[!h]
    \centerline{ \includegraphics[width=0.47\textwidth]{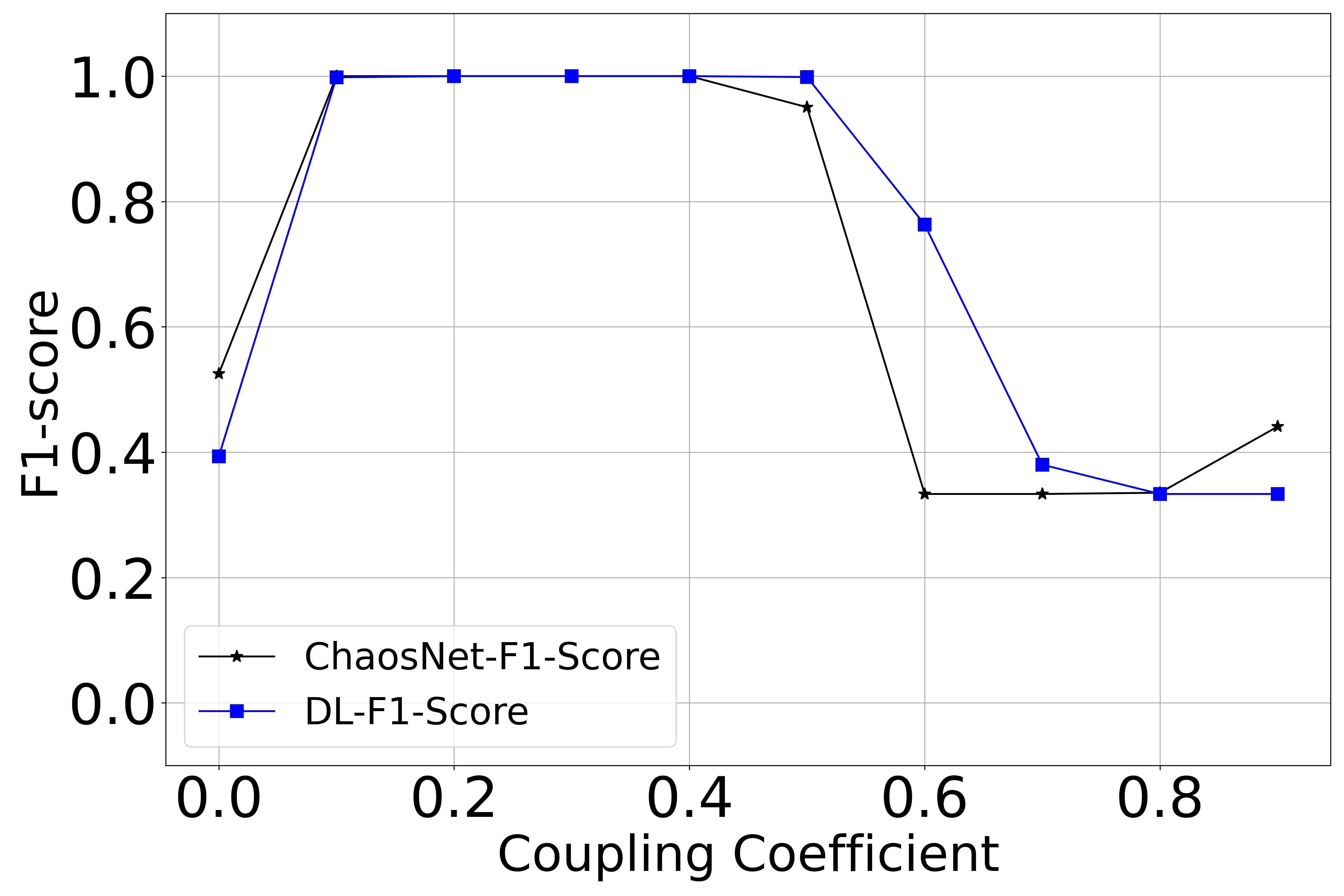}}
    \caption{Transfer Learning for Case IV: comparative performance of ChaosNet and five layer DL evaluated using macro F1-score for $\eta$ in the range $0$ to $0.9$ with a stepsize of $0.1$.}
    \label{Fig_Case_V_manifold_learning_logistic_map}
    \end{figure}
    
    \begin{figure}[!h]
    \centerline{ \includegraphics[width=0.47\textwidth]{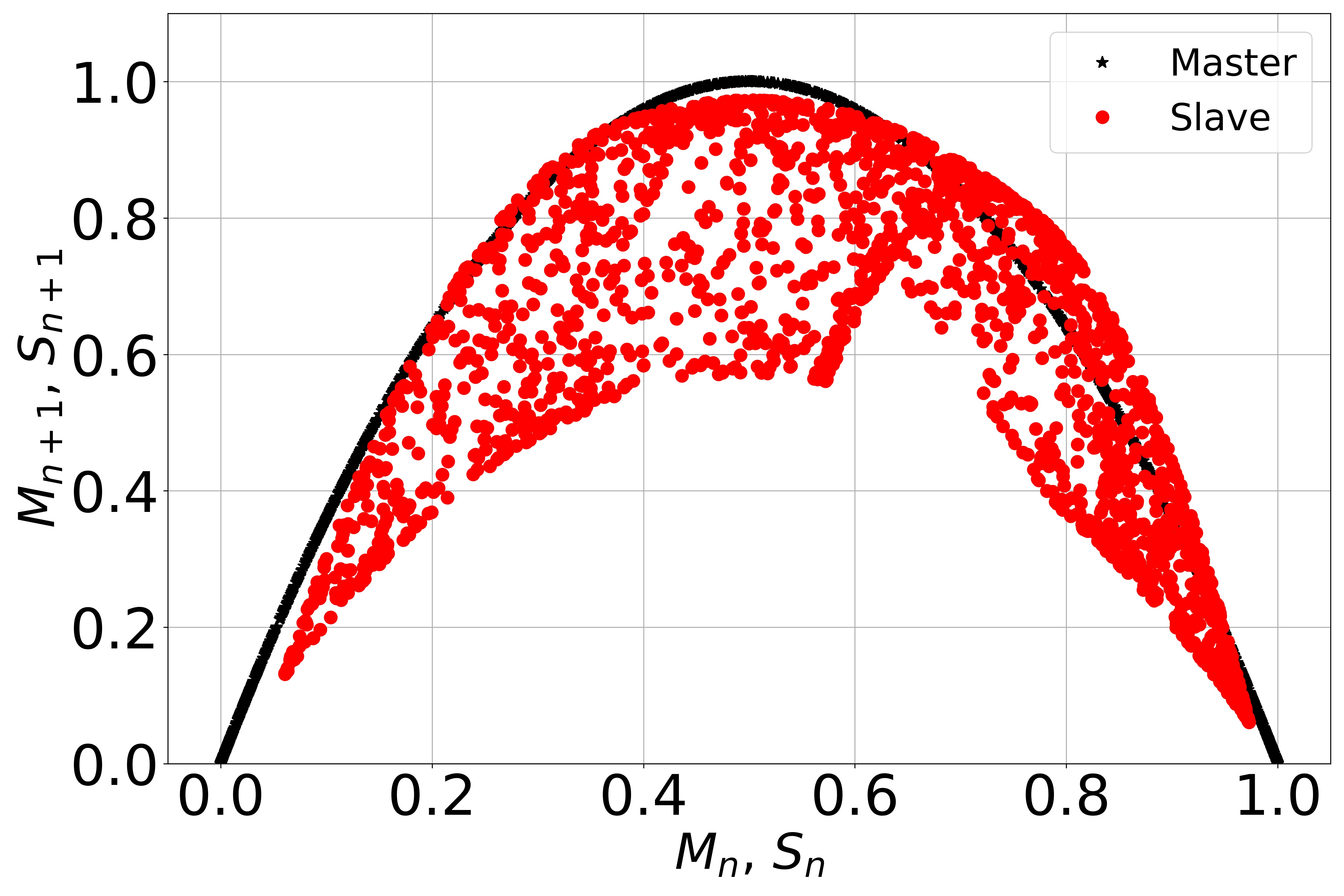}}
    \caption{Case IV: Attractor for the coupled 1D logistic maps in master slave configuration with $A_1 = 4.0$ and $A_2 = 3.82$.}
    \label{Fig_Case_V_attractor_logistic_map}
    \end{figure}

\end{itemize}


In the case of testing with data generated from different models, \verb+ChaosNet+ completely outperforms DL for Case I (Figure~\ref{Fig_Case_I_manifold_learning}) and Case II (Figure~\ref{Fig_Case_II_manifold_learning}) for the entire range of $\eta$. For Case III (Figure~\ref{Fig_Case_III_manifold_learning}) and IV (Figure~\ref{Fig_Case_V_manifold_learning_logistic_map}), \verb+ChaosNet+ and DL shows similar trends (with DL outperforming \verb+ChaosNet+ for some values of $\eta$). 
A high performance of \verb+ChaosNet+ in classification shows the separability of the mean representation vectors of cause and effect. 
\newpage
\subsection{Real Data}
\begin{figure}[!h]
    \centerline{ \includegraphics[width=0.47\textwidth]{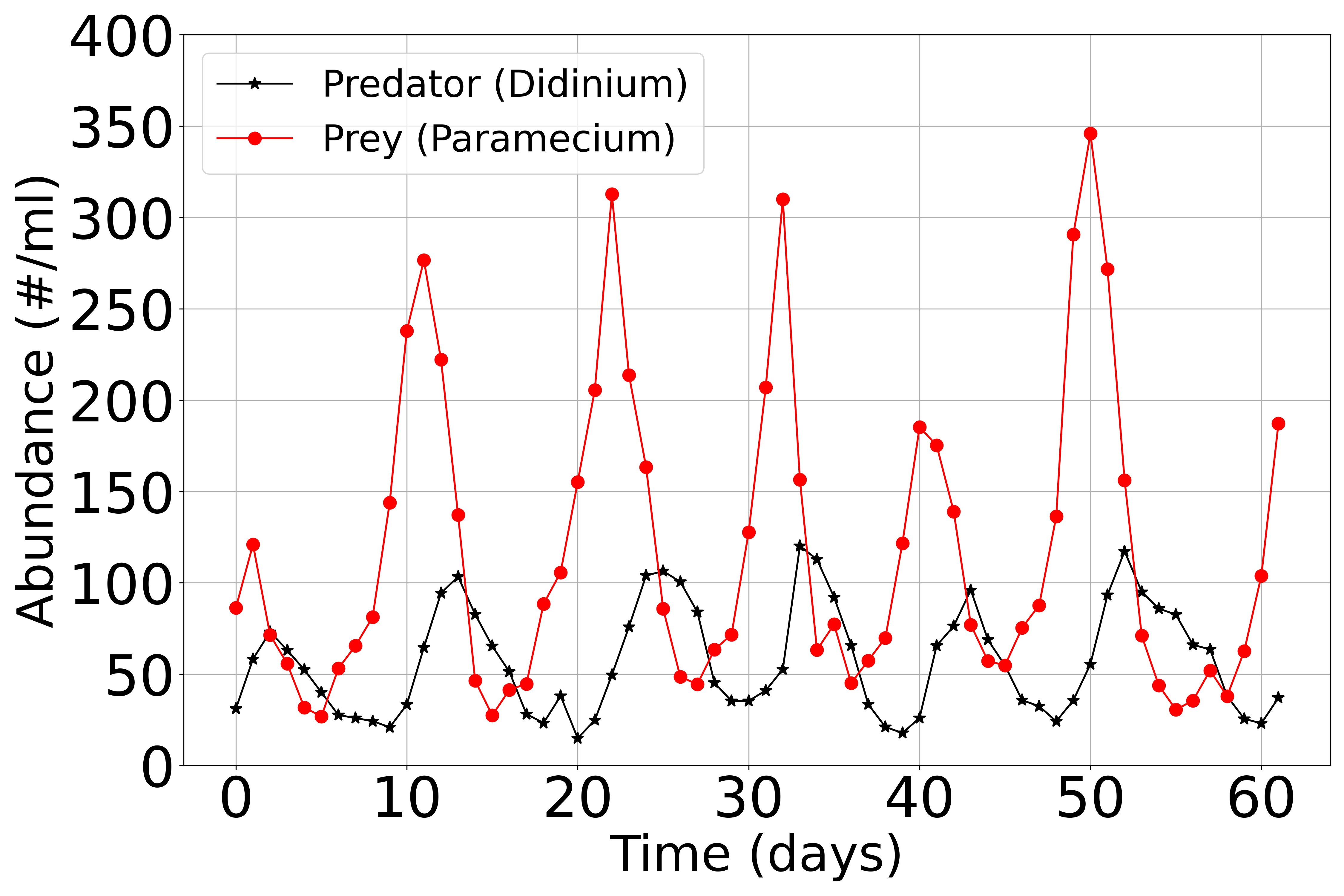}}
    \caption{Population dynamics of the predator (Didinium nasutum) and prey (Paramecium aurelia) as mentioned in ~\cite{veilleux1976analysis, jost2000testing}.} 
    \label{Fig_pouplation_dynamic_prey_predator}
    \end{figure}

The efficacy of ChaosFEX features in cause-effect preservation was evaluated on a real world dataset from a prey-predator system as well. The data consists of $71$ data points of predator (Didinium nasutum) and prey (Paramecium aurelia) populations~\cite{veilleux1976analysis, jost2000testing} (Figure~\ref{Fig_pouplation_dynamic_prey_predator}). This is a system of bidirectional causation as the predator population directly influences the prey population and then itself gets influenced by a change in the prey population. It is expected that the direct causal influence from the predator to the prey should be higher than in the opposite direction.

For our analysis, initial $9$ transients were removed. With the remaining $62$ data points, CCC values are computed for the raw data and ChaosFEX feature (firing time). The parameters of CCC\footnote{These were chosen using the selection criteria described in~\cite{kathpalia2019data}.} used for the raw data are $L=40, w=15, \delta=4, B=8$. In the case of ChaosFEX, firing time feature was extracted for the following NL hyperparameters: $q = 0.56$, $b = 0.499$, and $\epsilon = 0.1$. The CCC parameters chosen for ChaosFEX are $L=40, w=15, \delta=4, B=4$. The results for the cause-effect preservation for the raw data and ChaosFEX firing time feature is provided in Table~\ref{table_prey-predator_results}. CCC rightly captures the higher causal influence from predator to prey population and finds a lower influence in the opposite direction, for both raw data and ChaosFEX feature - firing time. 
\begin{table}[!h]
\centering
\caption{Cause-effect preservation of the prey-predator real world data using CCC.\label{table_prey-predator_results}}
\begin{tabular}{|c|c|c|}
\hline
Class   & CCC (raw data) & CCC (firing time) \\ \hline
$Predator \rightarrow Prey$ & 0.1160       & 0.0484     \\ \hline
$Prey \rightarrow Predator$ & -0.0210      & 0.0050 \\ \hline
\end{tabular}
\end{table}
%


\section{Limitations}
\begin{figure}[!h]
    \centerline{ \includegraphics[width=0.47\textwidth]{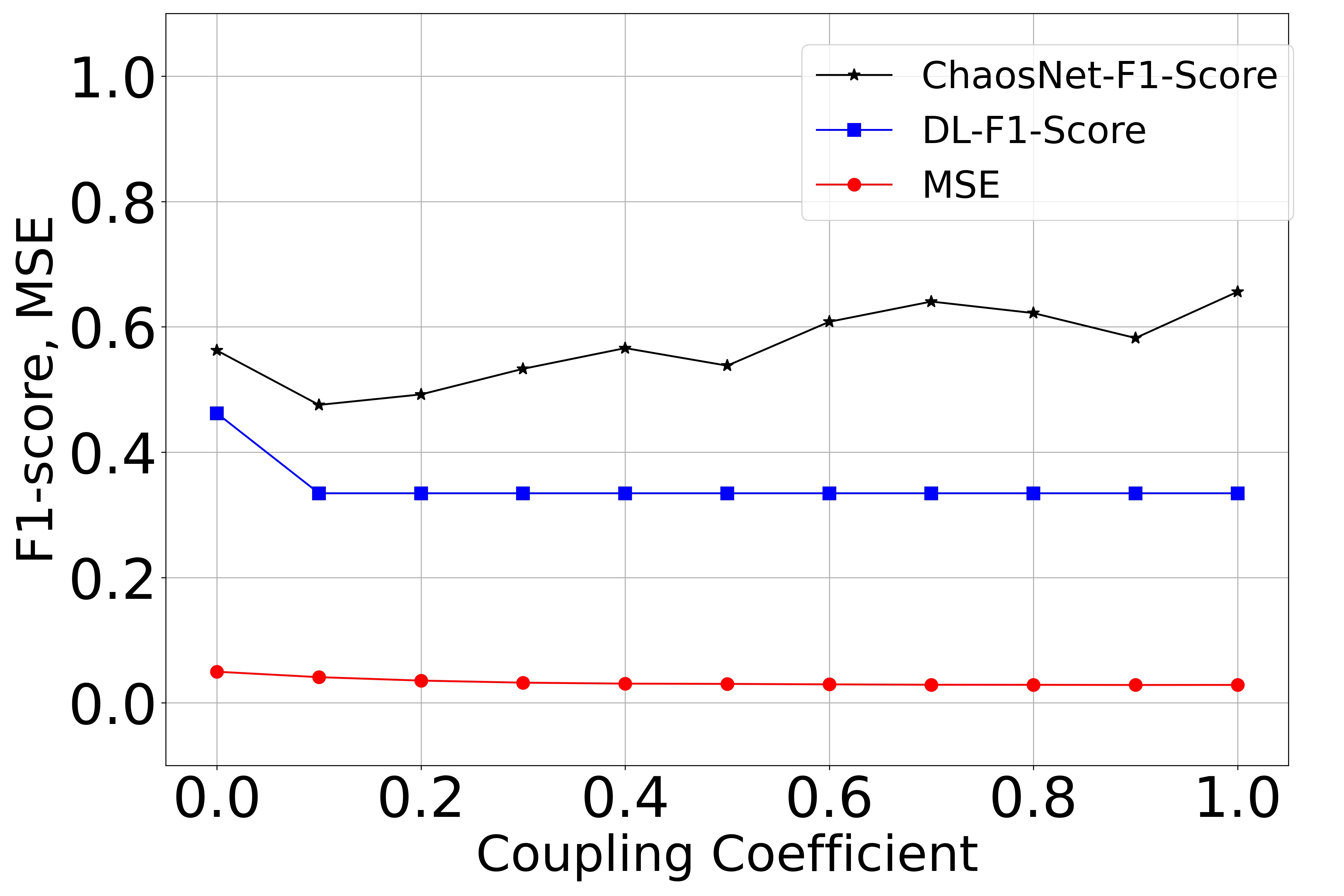}}
    \caption{F1-score vs. coupling coefficient for the classification of the coupled AR processes using ChaosNet.}
    \label{Fig_AR_chaosfex_classification}
    \end{figure}

In the case of coupled 1D chaotic maps, NL consistently performed well up to $\eta = 0.5$ for classification. However, the same is not true for the classification of data generated from coupled AR processes.  For $q = 0.78$, $b = 0.499$, and $\epsilon = 0.171$, the classification results are depicted in Figure~\ref{Fig_AR_chaosfex_classification}. In the same figure, it can be seen that DL performance is worse than NL\footnote{We have performed some amount of hyperparmater tuning for DL architecture, however a more extensive tuning needs to be performed.}. We have used the exact same architecture for DL as we have used for the cause-effect classification of data from coupled chaotic skew-tent maps in master-slave configure (section 4.2). 

A maximum macro F1-score = 0.656 was obtained for $\eta = 1.0$ implying that \verb+ChaosNet+ was not able to find mean representation vectors that could separate the two classes. Choosing a more sophisticated classifier for NL (instead of the simplistic cosine-similarity metric) could solve this problem and improve classification results. We shall explore these possibilities in a future study. 

In this research, we have shown the classification and preservation of causality for unidirectional causation of two variables. A detailed study needs to be undertaken for the classification and causal discovery for coupled high dimensional systems and real world datasets in the future.
\section{Conclusion}
In this work, Neurochaos Learning architecture - \verb+ChaosNet+ has been used in the classification of cause-effect for data generated from coupled chaotic maps. \verb+ChaosNet+ outperforms a five layer deep learning architecture in several cases. In the experiments, \verb+ChaosNet+ consistently outperformed a five layer DL upto a coupling coefficient of 0.7.  Causality testing using Granger Causality (for coupled AR processes) and Compression-Complexity Causality (for coupled chaotic systems and for a real-world prey-predator system) on the firing times extracted from the chaotic neural traces reveals the preservation of cause-effect in the NL feature extracted space. Further, the efficacy of the proposed method was observed in {\it transfer learning} of the classification of cause-effect from the master-slave time series generated from different chaotic unimodal maps (skew-tent maps with different skews and logistic map with different parameters). This motivates future research direction of NL in lifelong learning framework and classification of cause-effect using \verb+ChaosNet+ on real world datasets.

The preservation of causality can be attributed to the rich properties of the nonlinear chaotic transformation of GLS neurons in NL (\verb+ChaosNet+). Unlike traditional ANNs, NL is intrinsically a nonlinear deterministic algorithm that performs a point-by-point chaotic transformation, in fact, a non-linear embedding of the input raw features in to a high dimensional space.  {\it Deterministic Chaos} combines the best of both the worlds - {\it pseudo-randomness} and {\it determinism}. The ergodic, `random-like' structure of the chaotic neural traces enables an effective transformation of the input data (stimuli) preserving causality that is inherent in the input space and at the same time ensuring separability in the chaotic feature space for efficient classification.

\section*{Acknowledgment}
Harikrishnan N. B. thanks ``The University of Trans-Disciplinary Health Sciences and Technology (TDU)'' for permitting this research as part of the PhD programme. The authors gratefully acknowledge the financial support of Tata Trusts. A. Kathpalia acknowledges the financial support of the Czech Science Foundation, Project No.~GA19-16066S and the Czech Academy of Sciences, Praemium Academiae awarded to M. Palu\v{s}. Nithin Nagaraj gratefully acknowledge the financial support of Cog. Sci. Res. Initiative (CSRI), Dept. of Sci. \& Tech., Govt. of India under Grant No. DST/CSRI/2017/54(G).
%


\end{document}